\documentclass[letterpaper, 10 pt, conference]{ieeeconf}
\IEEEoverridecommandlockouts
\usepackage{graphicx}
\usepackage{float}
\usepackage{cite}
\usepackage{xcolor}
\usepackage{caption}
\usepackage{siunitx}
\usepackage{amsmath} 
\usepackage[bookmarks=true]{hyperref}
\newcommand{\robotname}{{{TerraSkipper}}}
\usepackage{todonotes}
\usepackage{booktabs}

\begin{document}
\title{\robotname: A Centimeter-Scale Robot for Multi-Terrain Skipping and Crawling} %


\author{Shashwat Singh$^{1}$, Sheri Zhang$^{1,2}$, Spencer Matonis$^{3}$, and Zeynep Temel$^{1}$%
\thanks{This work is supported in part by National Science Foundation grants 2308654.}%
\thanks{$^{1}$The Robotics Institute, Carnegie Mellon University, Pittsburgh, PA 15213, USA.}%
\thanks{$^{2}$Department of Mechanical Engineering, Carnegie Mellon University, Pittsburgh, PA 15213, USA.}%
\thanks{$^{3}$Edulis Therapeutics, Nashville, TN 37204.}%
\thanks{Corresponding author: shashwa3@andrew.cmu.edu.}%
}

\maketitle
\begin{abstract}

Mudskippers are unique amphibious fish capable of locomotion in diverse environments, including terrestrial surfaces, aquatic habitats, and highly viscous substrates such as mud. This versatile locomotion is largely enabled by their powerful tail, which stores and rapidly releases energy to produce impulsive jumps. Inspired by this biological mechanism, we present the design and development of a multi-terrain centimeter-scale skipping and crawling robot. The robot is predominantly 3D printed and features onboard sensing, computation, and power. It is equipped with two side fins for crawling, each integrated with a hall effect sensor for gait control, while a rotary springtail driven by a \SI{10}{\milli\meter} planetary gear motor enables continuous impulsive skipping across a range of substrates to achieve multi-terrain locomotion. We modeled and experimentally characterized the tail, identifying an optimal length of \SI{25}{\milli\meter} that maximizes the mean propulsive force (\SI{4}{\newton}, peaks up to \SI{6}{\newton}) for forward motion. In addition, we evaluated skipping on substrates where fin based crawling alone fails, and varied the moisture content of uniform sand and bentonite clay powder to compare skipping with crawling. Skipping consistently produced higher mean velocities than crawling, particularly on viscous and granular media. Finally, outdoor tests on grass, loose sand, and hard ground confirmed that combining skipping on entangling and granular terrain with crawling on firm ground extends the operational range of the robot in real-world environments.


\end{abstract}



\section{Introduction}

Robots traversing natural environments face significant challenges when encountered with soft, heterogeneous, and deformable terrains such as mud, sand, clay, and grass~\cite{li2013terradynamics, maladen2011mechanical}. Unlike rigid or artificial surfaces, these substrates vary widely in water content, compaction, and particle size, producing unpredictable mechanical responses under load. Locomotion mechanisms that are effective on flat or firm ground often fail in these conditions, particularly on the centimeter-scale~\cite{fearing2006challenges, stpierre2019toward, singh2024buffalo,haldane2013animal,birkmeyer2009dash,demario2018development} where robots cannot rely on heavy actuation~\cite{saranli2001rhex} or large appendages~\cite{baines2022multi}. Overcoming this challenge is essential for the deployment of small robots in real-world applications such as environmental monitoring~\cite{dunbabin2012robots}, wetland exploration~\cite{baines2022multi}, and search and rescue operations in disaster zones~\cite{murphy2017disaster ,vaquero2024eels, boateng2024heterogenous}.  

\begin{figure}
  \centering
  \includegraphics[width=0.8\linewidth]{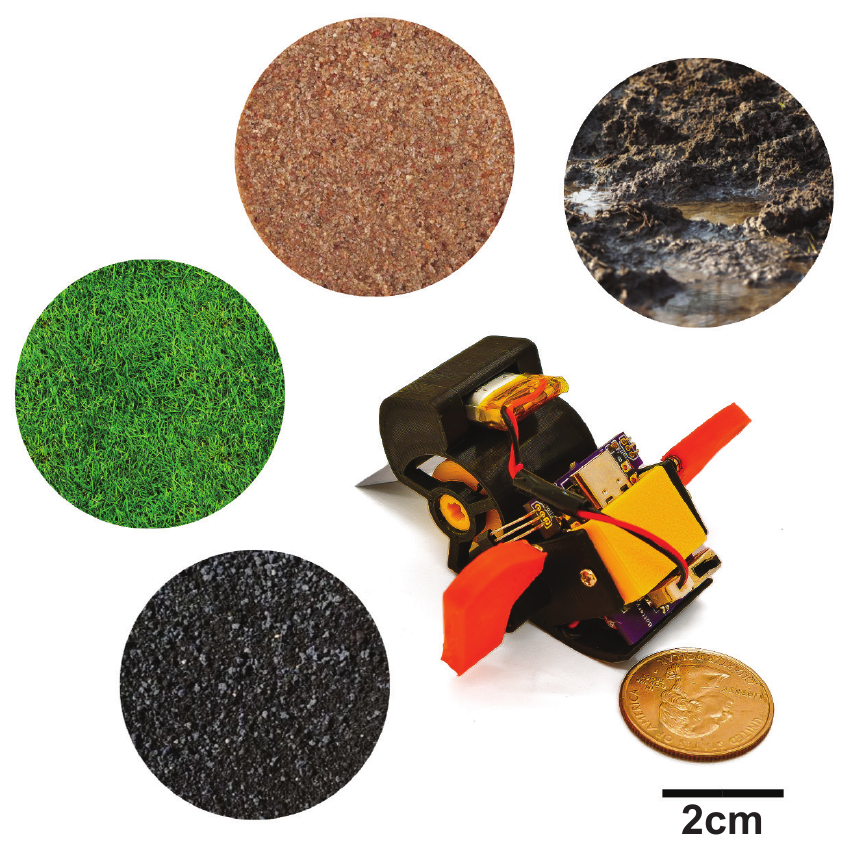}
  \caption{\robotname{} with representative challenging terrains shown in the surrounding circular pictures, on which the robot can locomote. A U.S.\ quarter and a ruler indicate scale.}
  \label{fig:robot}
\end{figure}

Biological systems provide valuable inspiration for designing robots capable of locomotion in complex environments~\cite{bledt2018cheetah, baines2022multi, vaquero2024eels}. Robots like mudskipper robot~\cite{lin2023mudskipper}, salamander robot~\cite{crespi2013salamandra}, and small fish robots~\cite{chen2018toward} exploit impulsive tail strikes and fin movements to propel themselves across semi-aquatic terrains where conventional crawling or swimming alone would be ineffective \cite{kawano2013propulsive, pace2009mudskipper}. Previous robotic studies have demonstrated the utility of tails for self-righting and inertial reorientation~\cite{johnson2012tail, libby2016comparative}, jumping~\cite{singh2024multi, ramirez2025springtail}, or propulsion on solid ground~\cite{yang2023proprioception} and in fluids \cite{lin2023mudskipper, chen2018toward}. Similarly, robots equipped with legs or paddles have been used to navigate granular substrates and shallow water \cite{baines2022multi, li2013terradynamics}. However, these approaches often focus on isolated functions such as jumping on rigid surfaces~\cite{haldane2017repetitive} or paddling in fluids~\cite{sun2023embedded} and rarely address the combined challenge of mobility in environments with variable rheology and deformability. 

Granular and muddy substrates, in particular, have been studied extensively in the context of terradynamics, where resistive force theory and empirical models have revealed how particle size, packing fraction, and intrusion depth influence locomotion~\cite{li2013terradynamics ,gravish2018robotics, zhang2014effectiveness}. Yet these studies have largely considered either simple intruders or larger robots, leaving open questions about how centimeter-scale robots interact with real, messy substrates under varying water content. At small scales, substrate robot interactions are further complicated by surface tension effects, local compaction, and stick–slip dynamics~\cite{li2013terradynamics}, making it difficult to generalize from existing models. Consequently, there remains a lack of systematic studies that connect substrate properties to robotic performance in biologically inspired, tail driven locomotion.  

In this work, we introduce \textit{\robotname{}} (Fig.~\ref{fig:robot}), a centimeter-scale robot that integrates a rotary spring actuated tail with lightweight fin to achieve multi-modal locomotion. The robot is predominantly 3D-printed, fully untethered, and features onboard sensing, computation, and power with an average runtime of \SI{30}{\minute} detailed in section~\ref{design}. Leveraging micromagnets and hall effect sensors, the fins provide lightweight (\SI{280}{\milli\gram}) feedback for gait control, enabling effective closed-loop synchronization while reducing overall mass and power requirements. This weight reduction enables continuous impulsive tail actuation with a single low-voltage motor, optimized through the springtail's impact force characterization discussed in section~\ref{tail_characterization}. Pectoral fin gait patterns were designed and evaluated using overhead video tracking, with and without encoder feedback, to assess their ability to move in a straight line in section~\ref{gait}. The resulting robot consumes less than \SI{1}{\watt} of power, and can be fabricated for under \$$50$, providing an accessible platform for investigating amphibious locomotion at small scales.
We further present a comprehensive experimental characterization of \robotname{} across diverse substrates in section~\ref{var_substrate} along with uniform black aquarium sand and bentonite clay powder with controlled water content in section~\ref{moisture_effect}. By systematically varying water content and compaction, we capture transitions between dry, and wide range of moisture content conditions, and quantify locomotor performance in terms of mean velocity and standard deviations across three trials. These experiments constitute a systematic study of impulsive tail-substrate interactions for a centimeter-scale robot, and highlight scenarios where tail driven propulsion succeeds and fin based locomotion alone fails. Among the representative centimeter-scale robots compared in Table~\ref{tab:benchmark}, \robotname{} is the only one demonstrated locomotion on both rigid and deformable terrain. Beyond reporting performance metrics, this work contributes a reproducible methodological framework for testing robots on deformable terrains, bridging bio-inspired design and granular physics. Collectively, these results advance the design of substrate aware robots, and suggest pathways for small amphibious platforms capable of robust locomotion in challenging environments, and other unstructured natural terrains.

\begin{table}[t]
\centering
\caption{Comparison with representative centimeter-scale robots.}
\label{tab:benchmark}
\setlength{\tabcolsep}{3pt}
\renewcommand{\arraystretch}{1.0}
\scriptsize
\begin{tabular}{lccccc}
\toprule
 & Terra & Buffalo & MM3P~\cite{demario2018development} & DASH~\cite{birkmeyer2009dash} & Veloci \\
Robot & Skipper & Byte~\cite{singh2024buffalo} &  &  & RoACH~\cite{haldane2013animal} \\
\midrule
Size (cm)
& 5.8×2.7×3.2
& 2.9×3.0×2.1
& 4.7×4.7×3.0
& 10×4.5×3.0
& 10×6.5×4.2 \\

Mass (g)
& 28
& 11.4
& 14.5
& 16.2
& 29.1 \\

Speed (mm/s)
& 31
& 235.7
& 57
& 1500
& 2700 \\

Speed (BL/s)
& 0.54
& 8.1
& 1.2
& 15
& 27 \\

Terrain
& Rigid /
& Rigid
& Rigid
& Rigid
& Rigid \\
& Deformable
& 
& 
& 
& \\

\bottomrule
\end{tabular}
\end{table}

\begin{figure}[!ht]
  \centering
  \includegraphics[width=\linewidth]{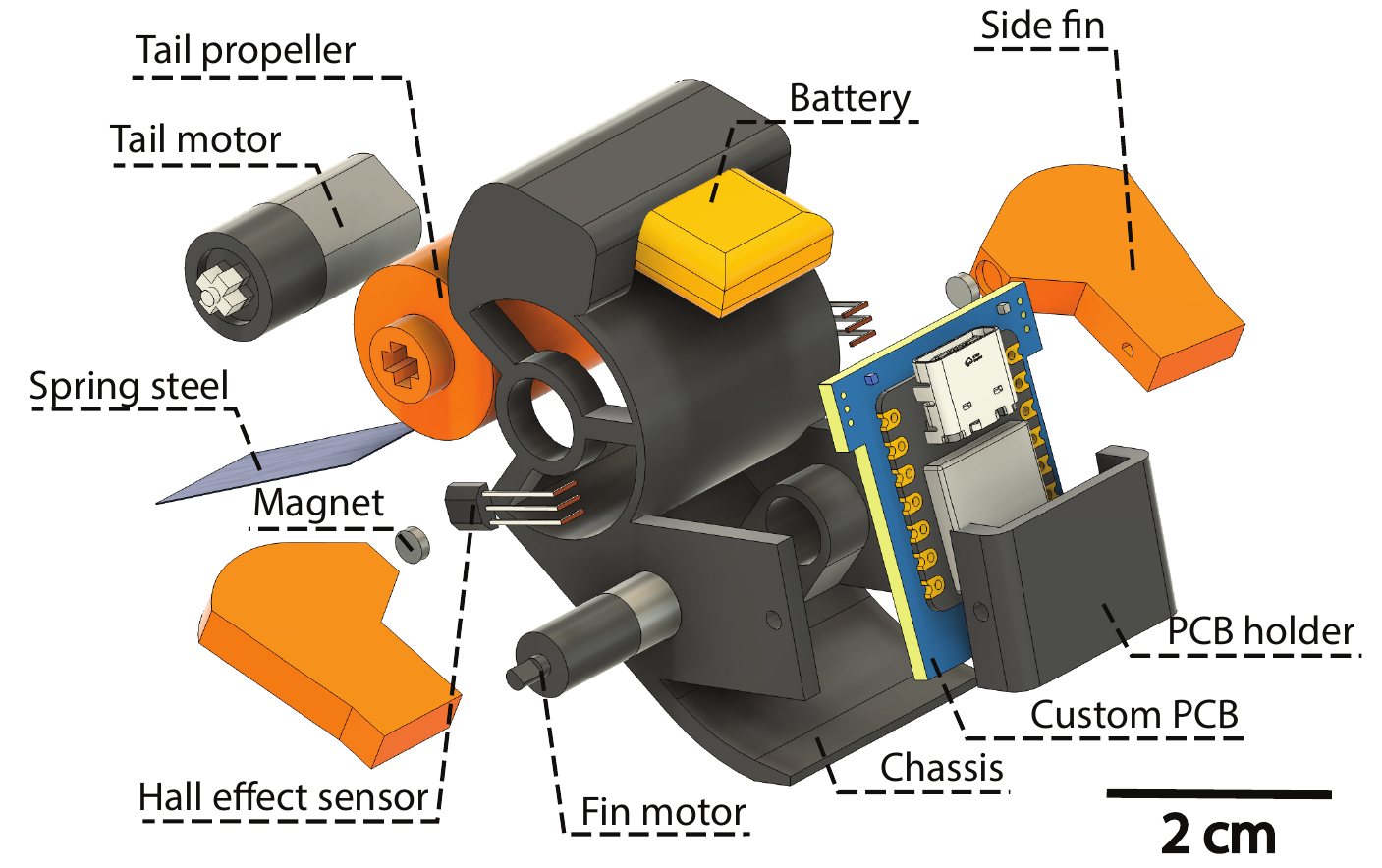}
  \caption{Exploded view of the CAD model of \robotname{}, highlighting key components: tail motor with propeller, spring steel element, hall effect sensor placement with embedded fin magnets, fin motors, chassis, battery, custom PCB, and PCB holder.}
  \label{fig:exploded}
\end{figure} 

\section{Design and Characterization}
\robotname{} is a centimeter-scale robot designed for autonomous operation in challenging terrains. It employs an impulsive spring actuated tail for skipping and soft fins with embedded magnets and hall effect sensors for coordinated gait control.  

\subsection{Robot Design}
\label{design}

\robotname{} was developed to satisfy three major requirements: (1) a palm-sized form factor, (2) locomotion across heterogeneous terrains, and (3) untethered operation with onboard power and control. To meet these goals, we designed a compact and lightweight platform (Fig.~\ref{fig:exploded}) measuring \SI{5.8}{\centi\meter} in length, \SI{2.7}{\centi\meter} in width (\SI{6.6}{\centi\meter} with fins), and \SI{3.2}{\centi\meter} in height, within a total mass of \SI{28}{\gram}. The design integrates two complementary modes of locomotion: springtail loaded skipping and fin driven crawling motion within a lightweight chassis to enable effective mobility on diverse substrates.

The chassis is fabricated using 3D-printed PLA (Bambu Lab X1-Carbon) and houses the custom PCB with electronics and sensor, actuators and battery. The fins are printed from TPU and embed two Neodymium magnets (SuperMagnetMan, D1018A, \SI{3.18}{\milli\meter} diameter, \SI{1.59}{\milli\meter} thickness) using super glue (Loctite 401), which interact with side mounted hall effect sensors (DRV5032, Texas Instruments) to provide feedback for gait control. The custom PCB hosts Xiao ESP32-S3 (Seeed Studio) dev board, which provides onboard computation and Wi-Fi-based communication for teleoperation. Two dual H-bridge motor drivers (DRV8833, Texas Instruments) supply power: one drives both fin actuators (Pololu, \SI{6}{\volt}, 700:1 Sub-Micro Plastic Planetary Gearmotors), while the other is dedicated to the tail actuator (E-S motor, RobotShop, 171:1, \SI{10}{\milli\meter} Planetary Gear Motor) to ensure sufficient current delivery by using both the driver output. Power is provided by a \SI{3.7}{\volt}, \SI{100}{\milli\ampere\hour} lithium-polymer battery, positioned at the rear of the chassis to maintain an optimal center of gravity during tail driven jumps. To improve landing stability, the underside of the chassis is curved upward at \SI{45}{\degree} in front of the fins, reducing the likelihood of overturning when the tail generates large impulses.  

The tail mechanism consists of a \SI{3}{\volt} DC motor (60~rpm) housed inside a propeller assembly. The motor is rigidly mounted to the chassis, while the propeller rotates freely around it, ensuring compact packaging without increasing the robot’s width. This encapsulated design balances weight distribution between the two sides of the robot while eliminating additional gearboxes or linkages, thereby simplifying the mechanism. A \SI{0.1}{\milli\meter} thick, \SI{10}{\milli\meter} wide spring steel element (Blue tempered, 1095 Shim sheet) is bonded using super glue (Loctite 401) into a slot within the propeller such that it remains perpendicular to the tangential face, ensuring full energy release during each skipping motion. The length of the spring steel sheet was optimized through characterization explained in the next section~\ref{tail_characterization} to generate maximum impact force for forward propulsion.


\begin{figure}[!ht]
  \centering
  \includegraphics[width=1\linewidth]{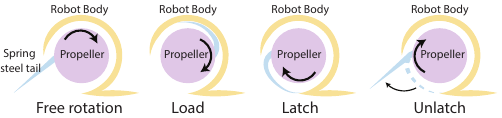}
  \caption{The four transition phases of the spring steel during the flicking motion of the jumper mechanism. Free Rotation: the spring rotates with the motor shaft without contacting the curved wall; Load: the spring tip engages the hollow cylinder and begins to bend, storing elastic energy; Latch: the spring is fully conformed and constrained inside the \SI{270}{\degree} arc, storing maximum energy; Unlatch: the spring tip exits the arc boundary, releasing stored energy as a rapid snapping motion that propels the robot forward.}
  \label{fig:model}
\end{figure} 

\subsection{Tail Characterization and Modeling}
\label{tail_characterization}

The springtail mechanism of \robotname{} is motivated by latch mediated spring systems~\cite{longo2019beyond} in biological and engineered jumpers, where elastic energy is gradually stored in a compliant element and then released in a rapid impulsive motion. In our design, the tail consists of a rectangular spring steel sheet (\SI{25}{\milli\meter} long, \SI{10}{\milli\meter} wide, and \SI{0.1}{\milli\meter} thick) mounted to a micro-planetary gear motor (E-S motor, RobotShop, 171:1, \SI{10}{\milli\meter} Planetary Gear Motor) that drives a rotating propeller hub. As the hub rotates, the springtail moves through four distinct phases of motion (Fig.~\ref{fig:model}). In the \textit{free rotation} phase, the spring rotates without contacting the chassis. During the \textit{load} phase, the tail tip engages with the curved housing, bending gradually and storing elastic strain energy. The \textit{latch} phase occurs when the spring conforms fully to the \SI{270}{\degree} arc of the housing, reaching maximum stored energy. Finally, in the \textit{unlatch} phase, the tail exits the arc boundary and snaps back to equilibrium, releasing the stored energy as an impulsive force that propels the robot forward.  

\begin{figure}[!ht]
  \centering
  \includegraphics[width=0.9\linewidth]{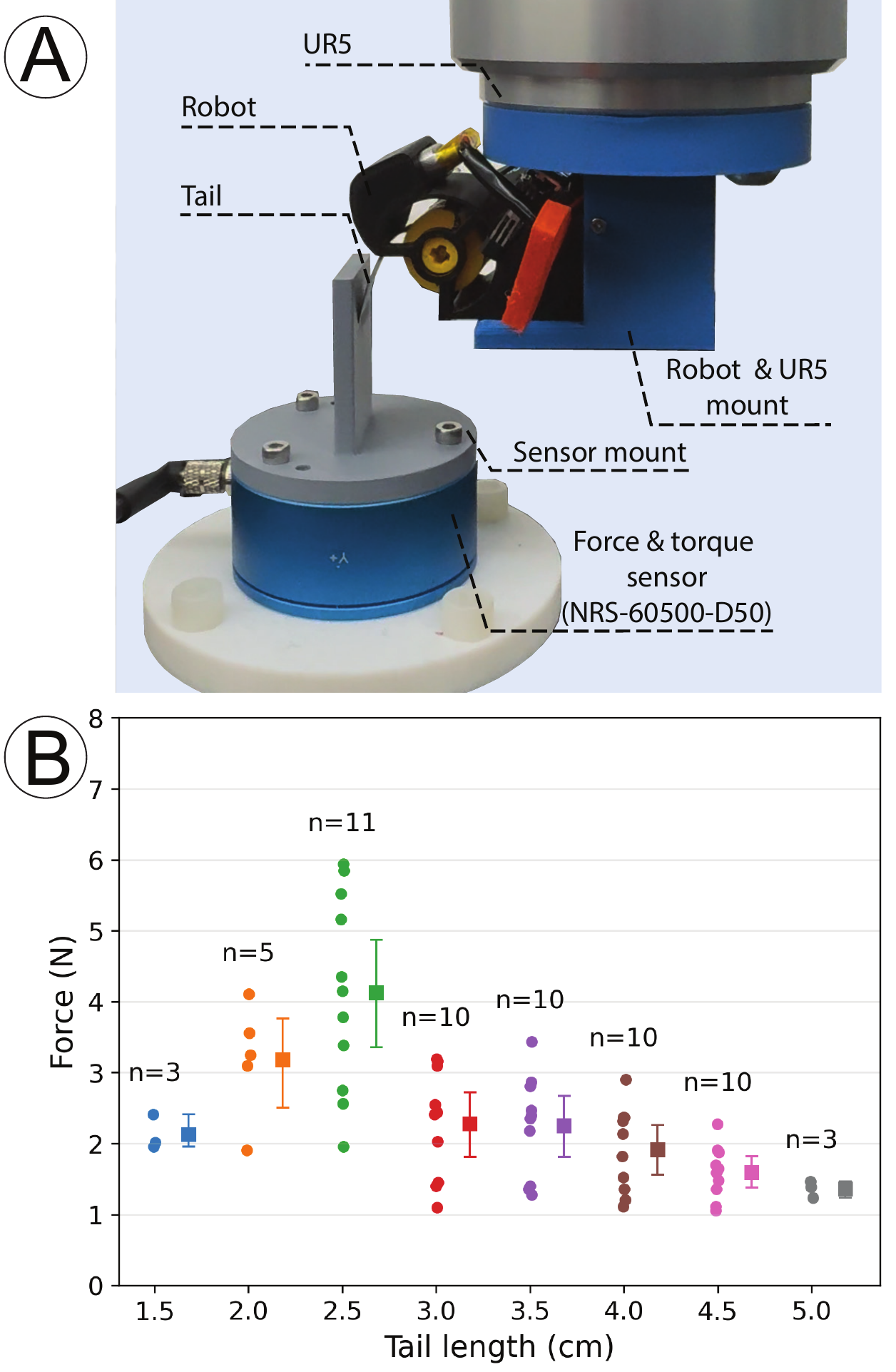}
  \caption{A) Experimental setup for tail characterization. The robot was mounted to the end effector of a UR5e robot arm via a 3D-printed mount. A force/torque sensor fixed on the table with an additional flat mount is used to record the tail’s impact force and frequency. B) Force measured by the force/torque sensor for different tail lengths. For each length, points show all peaks from \SI{10}{\second} recording; the square marker indicates the mean with a \SI{95}{\percent} bootstrap confidence interval. The annotation $n$ reports the number of peaks (tail strikes) detected within the \SI{10}{\second} window.}
  \label{fig:setup}
\end{figure}

\subsubsection{Tail characterization} 

To quantify the force output of the springtail mechanism, we conducted a series of tail characterization experiments for different tail lengths. The characterization setup and results are shown in the Fig.~\ref{fig:setup}. The robot was rigidly mounted to the end effector of a UR5e robot arm using a 3D-printed fixture, with the tail aligned against a flat surface mounted on top of a six-axis force/torque sensor (NRS-60500-D50). This setup in Fig.~\ref{fig:setup}-A allowed repeatable motor actuation sequences while measuring both the magnitude and frequency of tail strikes. Characterization data were used to optimize springtail dimensions and motor actuation speed, ensuring reliable unlatching and maximizing the impulsive forces generated during skipping locomotion.

For each configuration, all detected peaks from a \SI{10}{\second} recording are plotted in Fig.~\ref{fig:setup}-B as individual points, while square markers indicate the mean force with a \SI{95}{\percent} bootstrap confidence interval. The number of tail strikes $n$ within the recording window is also annotated. Across lengths, we observed a clear trade-off: longer tails produced more frequent strikes but at reduced peak forces, whereas shorter tails generated higher peak forces. Based on this trend, we selected a spring steel length of \SI{25}{\milli\meter} as the optimized configuration, as it consistently delivered mean force around \SI{4}{\newton} (with peaks up to \SI{6}{\newton}) while maintaining reliable and repeatable operation across trials. This balance of force output and consistency makes the \SI{25}{\milli\meter} tail the most effective choice to enable robust skipping locomotion.

The observed trade-off is further explained by geometric and actuation constraints: longer tails tend to roll within the hollow cylinder without striking the force/torque sensor, reducing measured peak forces, whereas shorter tails require higher motor torque to bend inside the housing and frequently jam, resulting in a low strike count ($n=3$).  

\subsubsection{Tail force model}
The springtail at unlatch can be approximated as a short cantilever of effective length $L_{\text{eff}}$ that conforms to the housing arc of radius $R$. For a rectangular strip of width $b$ and thickness $t$ with Young’s modulus $E$, the area moment is 
\[
I = \frac{b\,t^{3}}{12}.
\]
The equivalent tip stiffness of a cantilever of length $L_{\text{eff}}$ is
\[
k = \frac{3 E I}{L_{\text{eff}}^{3}}.
\]
When latched against the arc, the deflection is approximated by
\[
\delta \approx \frac{L_{\text{eff}}^{2}}{2R}.
\]
Combining $F \approx k\delta$ yields a compact expression for the peak elastic force at unlatch:
\[
\boxed{F \approx \frac{3 E I}{2 R L_{\text{eff}}}}, 
\quad \text{with } L_{\text{eff}} = R\theta,
\]
where $\theta$ is the engaged arc angle (in radians).

\paragraph{Intuition}
The model highlights that peak force grows rapidly as the engaged arc shortens: $F \propto 1/L_{\text{eff}}$ (or $F \propto 1/\theta$ at fixed $R$). Tail thickness also strongly influences force since $I \propto t^{3}$.

\paragraph{Numerical estimate for \robotname{}}
Using $E \approx \SI{200}{\giga\pascal}$ (1095 spring steel), $b = \SI{10}{\milli\meter}$, $t = \SI{0.10}{\milli\meter}$, and $R = \SI{11}{\milli\meter}$,
\[
I = 8.33\times10^{-16}\ \si{\meter^{4}}, 
\qquad \frac{3EI}{2R} \approx 2.27\times10^{-2}\ \si{\newton\meter}.
\]
The force can therefore be expressed as
\[
F~[\si{\newton}] \approx \frac{2.27\times10^{-2}}{L_{\text{eff}}~[\si{\meter}]}
\quad \text{or} \quad
F~[\si{\newton}] \approx \frac{2.06}{\theta~[\text{rad}]}.
\]

\begin{table}[!ht]
\centering
\begin{tabular}{c c c}
\toprule
Engaged angle $\theta$ & $L_{\text{eff}}$ (mm) & Predicted $F$ (N) \\
\midrule
$45^{\circ}$ ($\pi/4$ rad) & 8.64 & 2.6 \\
$30^{\circ}$ ($\pi/6$ rad) & 5.76 & 3.9 \\
$20^{\circ}$ ($0.35$ rad)  & 3.84 & 5.9 \\
\bottomrule
\end{tabular}
\caption{Predicted peak force values from the compact tail force model for different engaged arc angles.}
\label{tab:tail_force}
\end{table}

In practice, the engaged angle $\theta$ is not constant but varies considerably across strikes due to small differences in unlatching conditions, motor speed, and spring seating. Model predictions in Table~\ref{tab:tail_force} (for $\theta\!\in\![20^\circ,45^\circ]$) align with the measured \SIrange{2}{6}{\newton} range, supporting the interpretation that variability in engaged angle at unlatch is the primary source of force spread. Thus, while the compact model captures the scaling trend, the inherent inconsistency in $\theta$ contributes to the variation in experimental outcomes.

\subsection{Gait Control and Evaluation}
\label{gait}

Maintaining straight trajectories is a significant challenge for centimeter-scale robots, as small asymmetries in friction or actuation rapidly accumulate into drift~\cite{trenkwalder2019computational}. To mitigate this, we implemented a lightweight encoder by using a pair of hall effect sensor and a magnet embedded in the fin, to achieve controllable locomotion. Two gait patterns were implemented and evaluated: a \textit{synchronous gait}, in which both fins moved together and each cycle was validated by simultaneous sensor detection, and an \textit{asynchronous gait}, in which the fins moved alternately, with each waiting for the other to complete a rotation before proceeding. 

\begin{figure}[!ht]
  \centering
  \includegraphics[width=\linewidth]{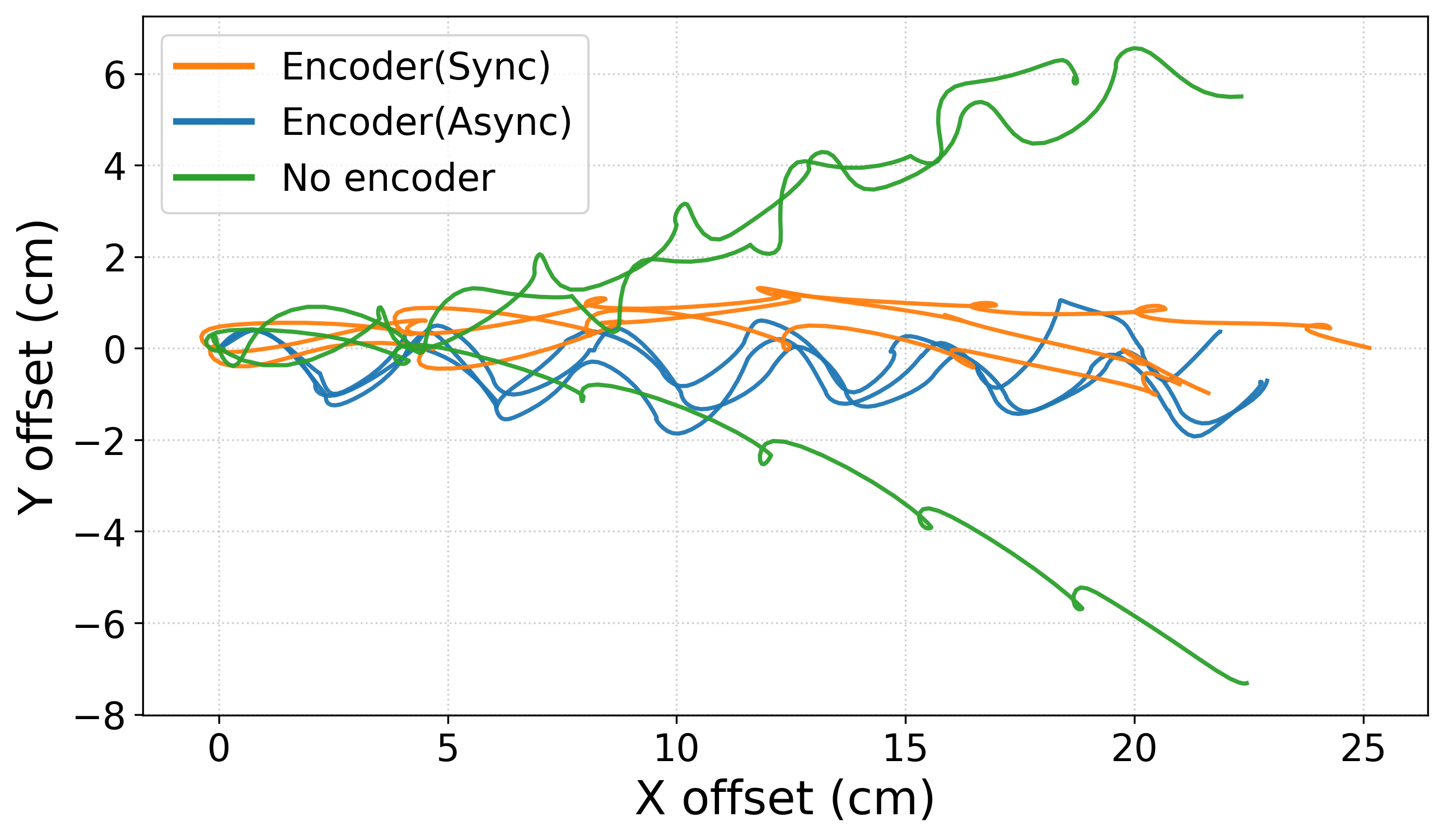}
  \caption{Top-down trajectories of \robotname{} tracked from overhead video on a smooth surface. For each condition, Encoder (Sync gait), Encoder (Async gait), and No encoder, three trials were performed and plotted to illustrate deviation from a straight path on a smooth acrylic surface.}
  \label{fig:control}
\end{figure}

For evaluation, robot trajectories were tracked from overhead video using tracking software (Tracker, physlets.org/tracker/) and plotted in the X–Y plane (Fig.~\ref{fig:control}). With encoder feedback, the robot maintained nearly straight trajectories in both gait conditions, deviating by under \SI{1}{\centi\meter} from the centerline over three trials. In contrast, when encoder feedback was disabled, the robot exhibited substantially larger drift, with deviations of up to \SI{6}{\centi\meter} in both positive and negative $y$ directions. These results highlight the importance of closed-loop control by using encoder feedback for maintaining directional stability and consistent locomotion, regardless of gait choice.


\begin{figure}[!ht]
  \centering
  \includegraphics[width=0.7\linewidth]{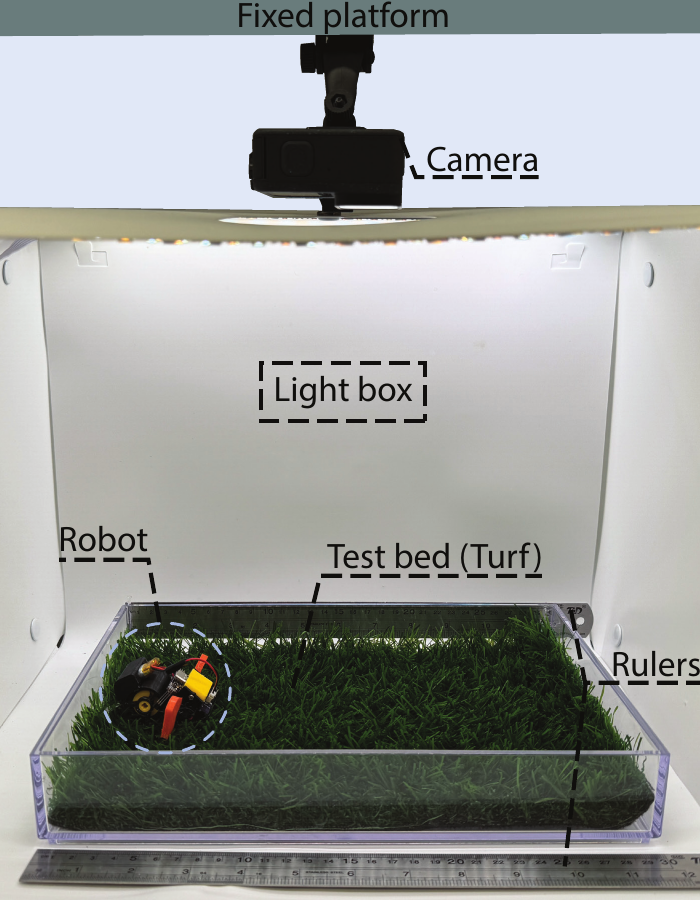}
  \caption{Experimental setup for tracking robot locomotion across various terrains. A clear container with different test beds was placed inside a light box, with rulers for scale calibration. A top-down video was recorded using a fixed overhead camera.}
  \label{fig:exp_setup}
\end{figure}

\section{Experimental Setup and Mobility Results}

We evaluated \robotname{} across a range of natural and artificial substrates to assess its locomotion performance under varying compliance, friction, and heterogeneity in Fig.~\ref{fig:substrate}. Substrates included bentonite clay mud, non-uniform desert beach sand, uniform black aquarium sand, artificial and natural grass, and flat rigid acrylic surfaces. It is essential to have such evaluations for understanding robot performance and enabling deployment of centimeter-scale robots in unstructured environments.

\subsection{Substrate Preparation and Tracking Setup}
Each substrate was prepared in a transparent acrylic container (\SI{20}{\centi\meter} wide, \SI{30}{\centi\meter} long, \SI{5}{\centi\meter} deep) placed inside a light box (Fig.~\ref{fig:exp_setup}). A GoPro Hero~12 Black mounted overhead on a fixed tripod provided top-down video for trajectory tracking, with rulers for scale calibration while tracking. All the substrates were leveled to a uniform depth of \SI{3}{\centi\meter} to ensure consistency across trials.

\subsection{Skipping on Various Substrates}
\label{var_substrate}
\begin{figure}[!ht]
  \centering
  \includegraphics[width=\linewidth]{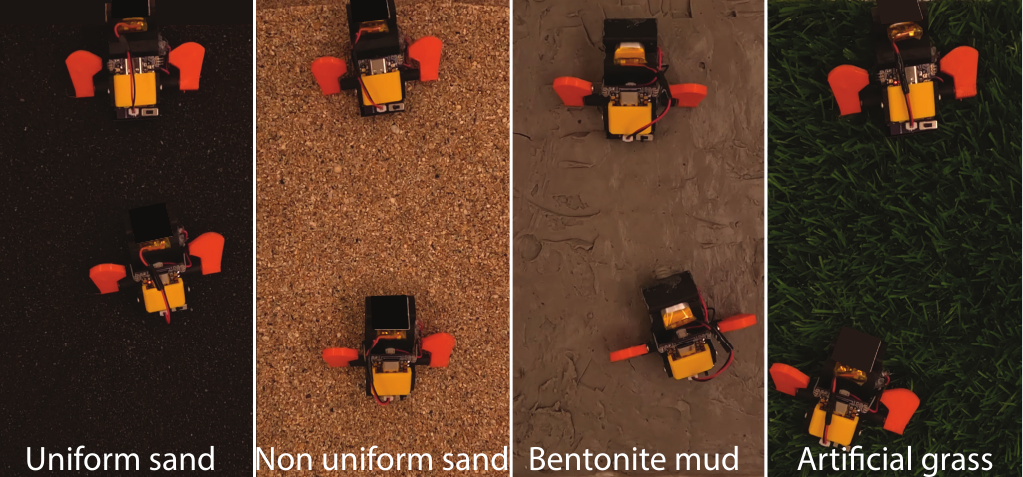}
  \caption{Snapshots of \robotname{} skipping on uniform aquarium sand, non uniform desert sand, bentonite clay powder mud, and artificial grass. These substrates highlight conditions where fin based crawling alone is ineffective or highly challenging.}
  \label{fig:substrate}
\end{figure}

We first evaluated skipping performance on substrates where fin based crawling is challenging such as sand, mud, and grass. Representative frames in Fig.~\ref{fig:substrate} show locomotion on bentonite clay powder (Pure original ingredients) mud with 1:3 water to clay powder ratio, non uniform desert sand (KolorScape, Paver sand step 2), uniform black aquarium sand (Imagitarium), and artificial grass (Grassclub). The trials were conducted with a fully charged battery, operating the robot at the maximum tail actuation frequency of \SI{1}{\hertz} or 60~rpm. Motion was recorded at 60~fps and analyzed frame-by-frame to extract displacement and velocity using tracking software (Tracker, physlets.org/tracker/).

\begin{table}[!t]
  \centering
  \begin{tabular}{lc}
    \toprule
    Substrate & Mean skipping speed (\si{\centi\meter\per\second}) \\
    \midrule
    Uniform aquarium sand   & $0.92 \pm 0.38$ \\
    Non-uniform sand        & $2.63 \pm 0.22$ \\
    Bentonite mud           & $1.24 \pm 0.12$ \\
    Artificial grass        & $5.38 \pm 0.71$ \\
    \bottomrule
  \end{tabular}
  \caption{Mean skipping velocities across different substrates with standard deviation for $n=3$ trials.}
  \label{tab:substrate_speeds}
\end{table}

Quantitatively, skipping produced mean speeds of $0.92 \pm 0.38$~\si{\centi\meter\per\second} on uniform aquarium sand, $2.63 \pm 0.22$~\si{\centi\meter\per\second} on non-uniform desert sand, $1.24 \pm 0.12$~\si{\centi\meter\per\second} on bentonite mud, and $5.38 \pm 0.71$~\si{\centi\meter\per\second} on artificial grass (Table~\ref{tab:substrate_speeds}). The ordering (grass $>$ non-uniform sand $>$ mud $>$ uniform sand) reflects how the substrate supports impulsive ground reaction: stiff grass and partially compacted sand transfer larger impulses, whereas loose granular beds such as uniform aquarium sand dissipate energy during tail impact.



\begin{figure}[!ht]
  \centering
  \includegraphics[width=\linewidth]{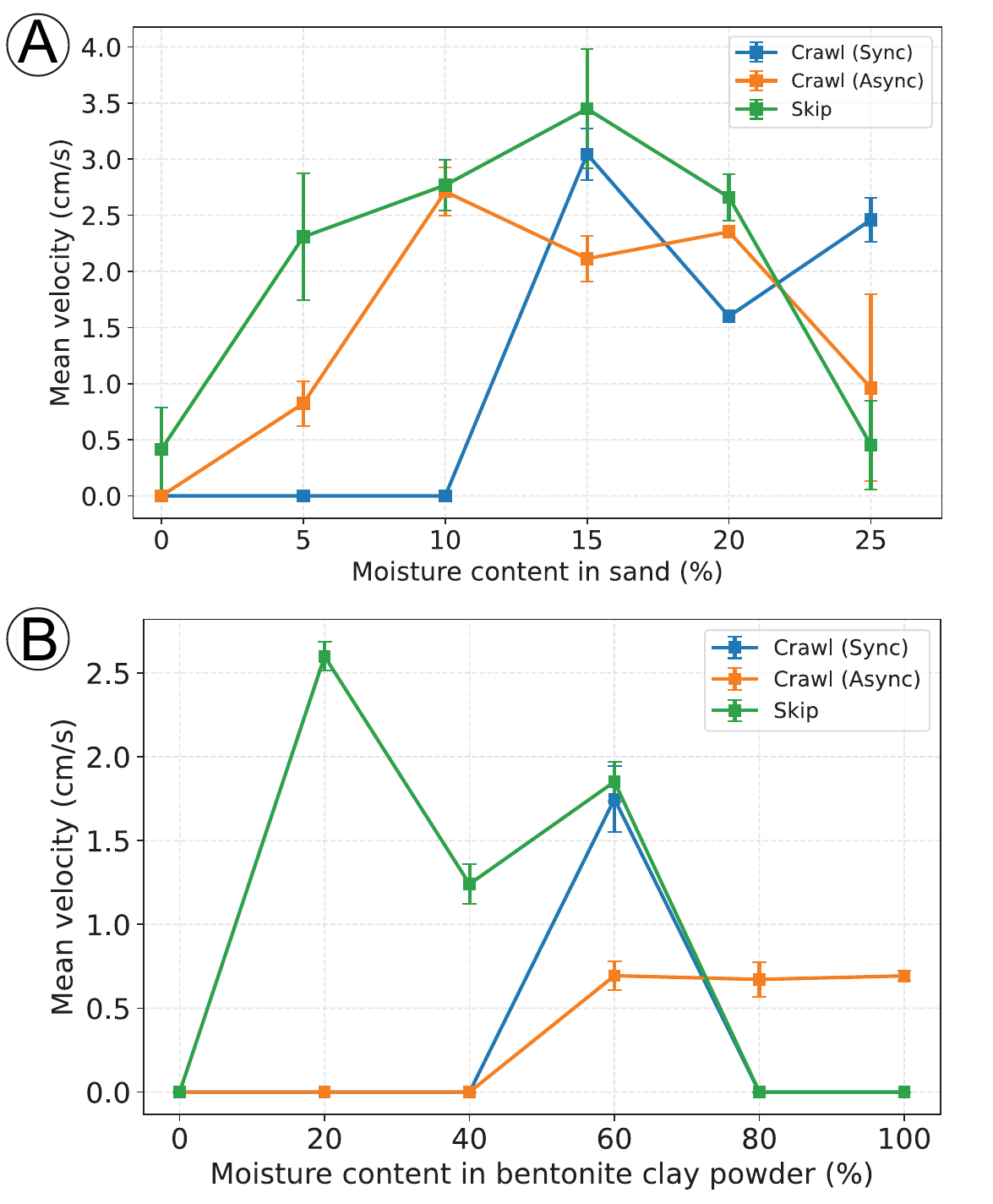}
  \caption{Mean velocity vs.\ moisture content for three locomotion modes on (A) aquarium sand and (B) bentonite clay. Moisture is percent water by mass relative to dry substrate. Each point is the mean of three trials; error bars show the standard deviations across trials. Trials with displacements $<\SI{10}{\centi\meter}$ were treated as failures.}
  \label{fig:moist}
\end{figure}

\begin{figure*}[!ht]
  \centering
  \includegraphics[width=0.85\linewidth]{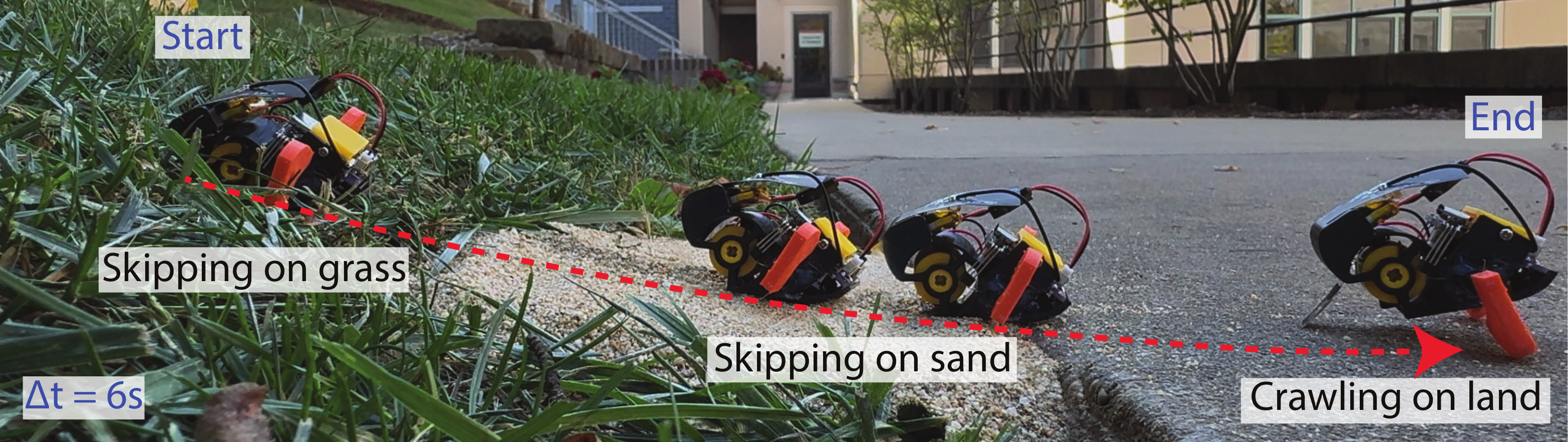}
  \caption{Side-view sequences of \robotname{} in outdoor tests. Panels show (left to right): skipping on grass, skipping on sand, and synchronous crawling on hard ground. Frames are overlaid every \SI{6}{\second}; each panel displays \SI{6}{\second} of motion.}
  \label{fig:demo}
\end{figure*}

\subsection{Effect of Moisture Content}
\label{moisture_effect}
In the previous section, we demonstrated the viability of skipping locomotion on diverse substrates where fin driven crawling alone is limited. To further investigate how locomotor performance depends on substrate properties, we systematically varied the water content in selected materials and compared the resulting crawling and skipping behaviors. Two substrates were chosen for this study: (1) uniform black aquarium sand (Imagitarium) and (2) bentonite clay powder (Pure natural ingredients). 

For sand, adding water alters the bulk mechanical response: dry sand behaves as a loose, frictional medium, while small amounts of water increase cohesion and improve footing, and higher contents lead to fluid-like behavior that reduces support. To prepare the test bed, we began with \SI{2}{\kilo\gram} of dry black aquarium sand (\SI{0}{\percent} water by mass) and incrementally added \SI{100}{\milli\liter} of water per iteration, hand-mixing for \SI{10}{\minute} to ensure uniformity, corresponding to a \SI{5}{\percent} increase in water content. At each level, three trials were performed for synchronous crawling, asynchronous crawling, and skipping, and the mean velocities with standard deviations were plotted as shown in Fig.~\ref{fig:moist}-A.

Fig.~\ref{fig:moist}-A shows how skipping followed a bell-shaped trend: velocity was lowest on dry sand (\SI{0.4}{\centi\meter\per\second}), peaked at \SI{3.4}{\centi\meter\per\second} at \SI{15}{\percent} water, and declined beyond this point as the substrate water content increased. Crawling required cohesive packing to be effective. Both modes failed in dry sand, but asynchronous crawling improved gradually at \SIrange{5}{10}{\percent} moisture to \SI{2.7}{\centi\meter\per\second}, while synchronous crawling reached its maximum of \SI{3.0}{\centi\meter\per\second} at \SI{15}{\percent}. Above \SI{20}{\percent}, both crawling gaits declined in speed but remained more effective than in the dry state. Overall, skipping was the better locomotion mode compared to crawling, achieving mobility across the widest range of moisture levels and delivering the highest peak velocity.

For bentonite clay, adding water similarly changes the substrate response but with higher absorption capacity than sand. To prepare the test bed, we used \SI{500}{\gram} of dry clay powder and added \SI{100}{\milli\liter} of water per iteration, hand mixing thoroughly for 10 minutes. This process allowed us to systematically increase the water content in increments while accounting for the material’s strong tendency to absorb and retain water. At each moisture level, three trials were performed for synchronous crawling, asynchronous crawling, and skipping, and the mean velocities were plotted (Fig.~\ref{fig:moist}-B).

Fig.~\ref{fig:moist}-B shows that skipping again outperformed crawling across different moisture content except at \SI{80}{\percent} and \SI{100}{\percent} moisture content because of the tail slippage. In the dry state, locomotion was negligible, but performance improved steadily with added water, reaching a peak of about \SI{2.6}{\centi\meter\per\second} at \SI{20}{\percent} moisture. Beyond this level, velocities declined as the clay transitioned from solid to a viscous slurry that dissipated energy and caused slip during locomotion. Crawling gaits followed a similar trend but at consistently lower velocities: asynchronous and synchronous crawling performed best around \SI{60}{\percent} moisture \SI{0.7}{\centi\meter\per\second} and \SI{1.7}{\centi\meter\per\second} respectively. Overall, skipping proved to be the most effective gait on clay, maintaining forward mobility across a broad range of water contents and achieving the highest peak velocity. 

Failures (i.e., points with a mean velocity of \SI{0}{\centi\meter\per\second}) were primarily due to (i) excavation and self-slippage into the substrate, which arrested forward motion, or (ii) pitching that lifted the forebody on firm ground and leaning on the tail, reducing contact and traction.

\subsection{Outdoor Demonstration}
We tested \robotname{} on natural outdoor terrains: grass, loose non uniform sand, and hard ground. Side-view time-lapse sequences are shown in Fig.~\ref{fig:demo}, with frames overlaid every \SI{6}{\second}. On grass and sand we used the skipping mode, because grass blades entangle the fins during crawling and the robot excavates into dry sand instead of advancing; consistent with section~\ref{moisture_effect}, synchronous and asynchronous crawling did not succeed on these substrates. In a continuous run, after \SI{12}{\second} the robot reached firm ground, where we switched to synchronous crawling for the next \SI{6}{\second}. On hard ground, tail–ground interaction imposes minimal drag and does not impede forward progression (see Supplemental Video). 

These results illustrate the benefit of switching locomotion modes in heterogeneous environments, impulsive skipping for compliant or entangling substrates, and crawling for firm terrain and highlight the effectiveness of the tail in natural settings. 

\section{Conclusion}
\label{sec:conclusions}

We presented \robotname{}, a centimeter-scale robot that blends fin based crawling (synchronous and asynchronous) with a springtail, impulsive skipping mechanism to traverse compliant and heterogeneous terrain with an average runtime of \SI{30}{\minute}. The fins are magnetically encoded with hall effect sensors to enable closed-loop gait timing; in contrast to open-loop operation. The feedback control yields stable straight-line trajectories with markedly reduced drift of less than \SI{1}{\centi\meter}. We also modeled and experimentally characterized the tail, identifying an optimal length that maximizes the averaged propulsive force of \SI{4}{\newton} with a maximum force of \SI{6}{\newton}.

We further conducted systematic substrate trials to evaluate how moisture content and compaction influence locomotor performance. Skipping consistently enabled forward motion in conditions where crawling alone was ineffective and achieved the highest peak velocities on moderately cohesive substrates. Crawling gaits became viable only once sufficient cohesion developed, with synchronous and asynchronous patterns showing complementary advantages. These findings establish skipping as the more robust mode of locomotion across variable substrates.

Beyond controlled laboratory tests, we validated the robot’s mobility in real-world environments. Outdoor demonstrations on natural grass, loose sand, and hard ground corroborated laboratory trends: skipping enabled progression on compliant or entangling substrates such as sand and grass where fin-driven crawling was ineffective, while synchronous crawling became advantageous on firm terrain. These results confirm that integrating impulsive tail actuation with fin-based control enables robust and versatile locomotion at small scales, and they motivate hybrid policies that adapt gait choice to substrate state.

Going forward, we see opportunities to (i) infer terrain class and slip from onboard inertial sensing (inertial measurement unit, IMU) to trigger closed-loop mode switching; (ii) develop algorithms that adapt the locomotion mode to local terrain rather than relying on teleoperation; and (iii) expand the field tests to evaluate design viability across diverse sites. These directions aim to broaden reliable centimeter-scale robot mobility and autonomy beyond the lab and into heterogeneous natural environments.

\bibliographystyle{ieeetr}
\bibliography{bibtex/bib}
\end{document}